\documentclass[sigconf,screen]{acmart}

\usepackage{multirow}

\AtBeginDocument{%
  \providecommand\BibTeX{{%
    \normalfont B\kern-0.5em{\scshape i\kern-0.25em b}\kern-0.8em\TeX}}}

\acmDOI{}
\acmISBN{}

\acmConference[ICAIF-NNAFA '23]{}{November 27,
  2023}{New York, NY}
\acmBooktitle{ICAIF '23: ACM International Conference on AI in Finance Workshop on NLP and Network Analysis,
 November 27, 2023, New York, NY} 

\setcopyright{none}
\begin{document}

\title{Improving Startup Success with Text Analysis}

\author{Emily Gavrilenko}
\affiliation{%
  \institution{California Polytechnic State University}
  \city{San Luis Obispo}
   \country{United States}
}
\email{eg6477@gmail.com}

\author{Foaad Khosmood}
\affiliation{%
  \institution{California Polytechnic State University}
  \city{San Luis Obispo}
   \country{United States}
}
\email{foaad@calpoly.edu}

\author{Mahdi Rastad}
\affiliation{%
  \institution{California Polytechnic State University}
  \city{San Luis Obispo}
   \country{United States}
}
\email{mrastad@calpoly.edu}

\author{Sadra Amiri Moghaddam}
\affiliation{%
  \institution{JP Morgan Chase}
  \city{San Jose}
   \country{United States}
}
\email{sadra.amiri@gmail.com}

\renewcommand{\shortauthors}{Gavrilenko, et al.}

\begin{abstract}
Investors are interested in predicting future success of startup companies, preferably using publicly available data which can be gathered using free online sources. Using public-only data has been shown to work, but there is still much room for improvement. Two of the best performing prediction experiments use 17 and 49 features respectively, mostly numeric and categorical in nature. In this paper, we significantly expand and diversify both the sources and the number of features (to 171) to achieve better prediction. Data collected from Crunchbase, the Google Search API, and Twitter (now X) are used to predict whether a company will raise a round of funding within a fixed time horizon. Much of the new features are textual and the Twitter subset include linguistic metrics such as measures of passive voice and parts-of-speech. A total of ten machine learning models are also evaluated for best performance. The adaptable model can be used to predict funding 1-5 years into the future, with a variable cutoff threshold to favor either precision or recall. Prediction with comparable assumptions generally achieves F scores above 0.730 which outperforms previous attempts in the literature (0.531), and does so with fewer examples. Furthermore, we find that the vast majority of the performance impact comes from the top 18 of 171 features which are mostly generic company observations, including the best performing individual feature which is the free-form text description of the company.

\end{abstract}






\maketitle

\newcommand{\totaldatapts}{22,125 }
\newcommand{\totalcomp}{9,842 }
\newcommand{\totalfunded}{6,574 }
\newcommand{\totalfundedperct}{29.7\% }

\newcommand{\anydatapts}{20,237 }
\newcommand{\anycomp}{9,380 }
\newcommand{\anyfunded}{6,052 }
\newcommand{\anyfundedperct}{29.9\%}

\newcommand{\adddatapts}{1,888 }
\newcommand{\addcomp}{1,438 }
\newcommand{\addfunded}{522 }
\newcommand{\addfundedperct}{27.6\%}

\newcommand{\totaltweets}{4,355,869 }
\newcommand{\totaltwitter}{13,951 }
\newcommand{\totalfoundtweets}{8,640 }

\newcommand{\traindatapts}{17,233 }
\newcommand{\testdatapts}{3,004 }
\section{Introduction}
Over 300 Million startups are created each year \cite{isp_startups_created}. Highly innovative, disruptive, and profitable startup companies are increasingly chased by investors, and accurately predicting success of these unicorns is extremely difficult due to the high level of uncertainty and risk associated with developing startups \cite{dellermann_2017}. Success can be measured in different ways, including revenue, merger and acquisition, and securing funding, with the last one being the dominant method amongst researchers \cite{dellermann_2017}. 5-10\% of startups typically fall under the successful classification, categorized by continuous growth and a growing user base. The ``unicorn milestone'', a company valuation of 1 billion dollars, is the pinnacle of success, coined in 2013 by venture capitalist Aileen Lee to describe the rarity of such ventures \cite{antosiuk_predicting_2021}. 


Recent developments in machine learning have made it possible to use data science to predict successful companies\cite{dellermann_2017}. Startup databases such as CrunchBase contain information on hundreds of thousands of startups that can be mined for organizational data, founder information, and funding rounds received \cite{antosiuk_predicting_2021}. However, while current machine learning approaches dominate in analyzing ``hard'' data, they fail to capture ``soft'' skills such as creativity and innovation crucial in leading a business to success \cite{dellermann_2017}.

Previous research has explored using solely CrunchBase data \cite{antosiuk_predicting_2021}, CrunchBase data and human computation \cite{dellermann_2017}, and finally CrunchBase data with web scraping to maximize the accuracy of machine learning models \cite{sharchilev_web-based_2018}. Recent studies have sought to utilize the ``wisdom of crowds'' to capture ``soft'' skills in organizations, with research showing that collective intelligence reduces the noise and biases of individual predictions \cite{dellermann_2017}. 


This paper investigates to what extent the ``wisdom of crowds'' can be captured through online, publicly available data and combined with hard, factual data to improve model performance. This includes social media sources, specifically Twitter, and online news articles collected through Crunchbase and the Google Search Engine. Tweets and news headlines referencing a company are collected and linguistic features are mined to uncover the author’s sentiment and the topic being discussed. These linguistic features, along with hard statistical data about the company collected from Crunchbase, are used to predict whether or not a company is classified as ``successful''.

Previous research in startup success prediction has focused on different metrics of success, ranging from an IPO, M\&A, founding round, or continuous operation for 5+ years. We use fundraising events as milestones to label a company as successful. An investor's main goal is to see a return on investment, and that is achieved when a company raises an additional funding round. Since very few companies go public or become acquired, we submit this fundraising metric will best capture a startup's success.  
\section{Related Work}

Considering the vast impact of startups and the huge potential for return on investment, there has been a long history of research on determining what factors are crucial for business success. Initially, researchers relied on questionnaires \cite{stuart_abetti_1987}, but soon moved on to machine learning using existing databases such as CrunchBase and TechCrunch \cite{xiang_2012}. Soon general web data was used to augment these models \cite{sharchilev_web-based_2018}\cite{unal_machine_2019}. 

Some recent studies found that considering only geographical, demographic, and general company information results in precision, recall, and F1 scores of 57\%, 34\%, and 43\% respectively \cite{antosiuk_predicting_2021}\cite{zbikowski_machine_2021}. This study had the largest training set to-date, consisting of 213,171 companies. It found that location and industry of a company are some of the key indicators of startup success. However, bias exists in most data sets especially due to disproportionate popularity of certain geographic locations such as the San Francisco Bay Area \cite{murray_overcoming_2021}.

The latest work on startup success prediction focused on using free, publicly available web information as their data source \cite{garkavenko_2022}. While previous work focused on using structured databases for their machine learning models, building and maintaining these databases requires a tremendous amount of human effort. Garkavenko and her team explored whether freely available data such as the website of a startup, its social media activity, and its web presence can be used to predict funding events within a given time horizon. They started off by gathering 22k startups from hubs, investors, and conferences around the world, primarily focusing on European startups. Then, they used the startup's own website to extract general information such as the country of origin, age, number of employees, and number of  offices. Next, they gathered social network data such as the amount of social media accounts the startup had, collected though links on the startup's website, and their activity on Twitter including number of tweets, likes, and received mentions by other users. Then, they summarized the financial history of the startup containing previous funding round, last fundraising amount, and time since last secured round. This data was extracted from tweets and news articles using regular expressions. They had a false positive rate of 8.5\% and a false negative rate of around 6\% for a small sampled data set of 200 startups. Finally, they used the free Google Search API to retrieve the top 10 results for each startup and gathered the number of relevant results that mentioned the company name, the number of total results, and the number of results from each of the 500 popular domains. They tested model performance using the most widely used machine learning models such as Logistic Regression, Random Forest, and a gradient boosting algorithm supporting categorical variables called CatBoost. 

\subsection{Twitter for Prediction}
Twitter data has often been used for machine learning prediction, for example to pinpoint victims and their locations during natural disasters \cite{twitter_disaster_classification}, or to predict the outcome of English Premier League football matches \cite{twitter_football}, and predicting election results and political candidates' likelihood of election \cite{twitter_spanish_elections_2012, twitter_elections_2021}. 

In their paper, Antretter et al. \cite{antretter_predicting_2019} relied solely on Twitter data to predict startup survival, whether or not a startup would be alive in another 5 years. They collected Twitter activity, measured through likes, followers, and the sentiment of user comments, for 253 seed or early-stage companies founded between 2006 and 2018, of which 72\% survived for at least five years. Using the Twitter REST API, they collected a total of 187,323 tweets, 102,501 retweets, and 441,583 likes to train the model. They achieved an recall of 86\% and a precision of 80\% in correctly classifying startup survival, indicating that startups classified as surviving have a probability of 83\% of actually surviving. While this work shows the importance of Twitter data in startup prediction, survival is rarely used as a success metric in practice. We build upon their findings to show social media's impact on predicting the more popular measurement of startup success: funding events. 

In a more recent study from 2021, Tumasjan and his team continued the work in exploring Twitter sentiment's correlation to startup success \cite{tumasjan_twitter_2021}.
The researchers found that while Twitter sentiment was able to predict valuations, it failed to show a significant relationship to investment success, classified in this paper as an IPO or acquisition of the firm. They found that investors across industries are paying more and more attention to Twitter sentiment, but that ``hype'' generated by the masses is often due to over-optimism that isn't the best indicator of long term success. On the other hand, they found that patents are a strong indicator for IPO or M\&A, and that the effect of Twitter sentiment is stronger if a startup has applied for patents.

\section{Methodology}
\begin{table*}
\caption{
    \label{tab:feature-overview}Startup Feature Overview
}

\begin{tabular}{ | c | l | p{3.2in} | l |} 
    \hline
    Category & Name & Description & Type \\
    \hline
    \hline
    General & Description & Company description & Text\\ 
        & Company age & Months since founding & Num.\\ 
        & Founder count & Number of founders & Num.\\ 
        & Industries & Distinct industry categories listed on Crunchbase & Num.\\ 
        & Name length & How long is the company's name & Num. \\
        & Website length & How long is the company's website url & Num. \\  
        & Social media & Twitter, Facebook, and LinkedIn accounts & Num. \\ 
        & Country & Location of company headquarters & Cat. \\  
        & State & Location of company headquarters & Cat. \\ 
        & Headquarter hub & Four fields for CA, NY, TX, or other hub HQ & Cat. \\ 
    \hline
    Funding & Num previous rounds & Total number of previous fundraising rounds & Num.\\  
        & Last fundraising amount & Money raised (USD)  during the last round & Num.\\ 
        & Months since fundraising & Months since last fundraising round & Num.\\  
        & Last fundraising stage & Angel, seed, series A, etc. & Cat. \\ 
        & Num total investments & Total number of investments received & Num.\\  
        & Num distinct investors & Number of distinct investors across all rounds & Num.\\ 
        & Distinct/total investors & 0-1 scale for capturing repeat investors & Num.\\ 
    \hline 
    News & Num articles & Total number of Crunchbase articles & Num.\\ 
        & Top publisher count & \# of articles written by a top 10/50 publisher & Num.\\ 
    \hline
    Google & Total results & Total Google Search results for the company name & Num.\\ 
        & Own company results & Number of times the company's website, LinkedIn, Twitter, or Facebook appeared in top 10 results & Num.\\
        & Top publisher count & \# of results by a top 10/50 publisher website & Num.\\ 
        & Top google count & \# of results by a top 10/50 Google website & Num.\\ 
    \hline  
    Twitter & Tweet count & Total number of tweets & Num. \\
        & Company count & Total number of tweets by company & Num. \\
        & Unique users & Total number of unique users tweeting & Num. \\
        & Engagement metrics & Likes, retweets, replies, quotes & Num. \\ 
        & Sentiment scores & VADER and Bert sentiment & Num. \\ 
        & Tweet contents & \% tweets containing company website, account, replies, mentions, hashtags, links, and emojis & Num. \\ 
        & Linguistic Features & Tweet structure: characters, words, sentences, shape; and deep linguistic: passive voice, parts-of-speech, syllables, reading score, complex words, etc. & Num. \\ 
    \hline
\end{tabular}
\end{table*}
In this section, we describe the process of building our dataset of \totalcomp companies, and training our prediction models using data extracted from Crunchbase, the Google Search API, and Twitter. 

\subsection{Company Selection}
\totalcomp companies are selected for this study, with \totalfundedperct of the data points constituting the positive class, classified by them raising a funding round in the allotted time window. The companies are randomly selected from Crunchbase or Pitchbook, with 2500 companies queried for having raised a funding round, while the rest are randomly chosen from the Crunchbase 1 million free-tier company dataset \cite{crunchbase_csv_export}. 

Each data point in our dataset consists of a company at a specific point in time. Since many of the companies are older, several data points are generated for the same company at different stages of development, however we never perform prediction on a point using training material from the future.

\subsection {Data Collection} \label{data_collection}
We collect data from Crunchbase, the Google Search Engine, and Twitter for generating our feature set. An overview of the data sources can be viewed in Table \ref{tab:feature-overview}.

\subsubsection {Crunchbase} \label{Crunchbase_Implementation}
Detailed information about each company is collected using the Crunchbase API using information from the organizations, press references, and funding rounds tables. The organizations table includes general information about companies such as name, headquarter address, number of employees, website, social media links, and status of the organization – active, closed, acquired, or ipo. The press references table includes all the news articles collected for a company, including the article title, publisher, and publication date. The funding\_rounds table includes information on all funding events, such as the announcement date, amount raised, and investment type (seed, angel funding, series A, B, C, etc.). 

\subsubsection {Google Search Engine}
Google Search results are collected for each entry consisting of the company and a point in time from which the prediction is made. The free tier Google Search API is used to fetch the top 10 results for every company. For every search, the company name is used as the search term, and a date range is used to limit look-ahead bias by filtering out search results unavailable at the time of prediction. The start date is set as one year before the company was founded to limit irrelevant results published years before the company was born. Each response returns the total number of results found by Google, and a link, title, and snippet for each one of the top 10 results. It's important to note that the total number of results contains a majority of irrelevant matches that Google returns, similar to the actual search engine behavior. 

\subsubsection {Twitter} \label{twitter_features}

\begin{table}
\caption{
    \label{tab:tweet-filters} Constructing Twitter Query
}
\begin{tabular}{ l | l } 
    \hline
    Tweet filters & Syntax\\
    \hline
    contains company website & website\_url \\ 
    posted by company & from:\textit{company\_username} \\ 
    repost of company tweet & url:\textit{company\_username} \\
    company mentions & @\textit{company\_username} \\
    excludes retweets & -is:retweet \\
\end{tabular}
\end{table}

All tweets that either contain the company website, are posted by the company Twitter account, are re-posts of a company tweet, or directly mention the company's twitter account are retrieved (Table \ref{tab:tweet-filters}). Initially, tweets were queried that also contained the company name, but those were eventually discarded due to the inability to distinguish false matches in the returned results. Additionally, tweets are filtered by date to only include those posted 9 months before the specified prediction date to avoid look-ahead bias. 

A total of \totaltweets tweets were retrieved for \totaldatapts distinct training and evaluation data points. \totaltwitter(63.1\%) of those data points have a company twitter account and \totalfoundtweets (39.1\%) have at least one tweet result returned for the provided 9 month period. This results in an average of 201 tweets per company, but the mean is highly influenced by the highly active accounts with over 1,000 collected tweets, 4.0\% of the total datapoints. A default value of 0 is used for all tweet features where no tweets are collected within the time horizon.  

\subsection {Feature Extraction} 
The features can be divided into five broad categories according to the information sources that they capture: general, news, funding, web search, and twitter. We describe each section in more detail below. For more about the feature extraction process and the generated feature lists, see \cite{gavrilenko_2022}.

\subsubsection{General Company Data}
General company information such as the company name, free-form text description, months since founding, number of founders, and website url are collected directly from the Crunchbase Search API. 

\subsubsection{News Articles} \label{news_articles_data}
The Crunchbase press\_references table is used to fetch the latest 2,000 news articles for a company. A total of 38,660 press references were retrieved for the \totalcomp companies in our dataset. This collection of press references is used to determine the top 10 and top 50 most popular publishers for news articles on Crunchbase. The top five publishers on Crunchbase are TechCrunch (1,714 occurrences), PR Newswire (849), Business Wire (753), PRNewswire (696), and PRWeb (528). It is inferred that a publisher who commonly posts about startups across industries and regions is a credible source on new ventures, and it is a high accomplishment to be featured on their site. The amount and percentage of press references written by a popular publisher are used to calculate the top\_10 and top\_50 publisher count features. 

\subsubsection {Funding Events} 
The Crunchbase funding\_rounds table is used to gather the number of previous fundraising rounds, previous fundraising amount, months since last fundraising, the last funding stage, number of total investments and distinct investors, and the ratio between distinct and total investors. 

\subsubsection {Google Search Results}
Given a company and prediction date pairing, the top 10 Google Search results and the total result count are returned as raw data. Only results available at the prediction date are fetched to avoid look-ahead bias. The top 10 result links are used to count how many times the company website or social media links (Twitter, Facebook, and LinkedIn) appear in the results. A high number of matching company links was hypothesized to correlate with funding as the company scores high in SEO. Additionally, the top 10 and top 50 websites appearing in Google Search results are calculated to be top referrers and used as a feature. 

\subsubsection {Twitter} \label{collected_tweet_features}
68 features are calculated for every tweet and included in the final prediction model. 

Public metrics are returned for every tweet and contain the (a) number of likes, (b) number of retweets, (c) number of replies, and (d) number of quotes per tweet. These metrics are aggregated for the nine-month period and the average count per tweet and total count over the time period are included in the prediction dataset. 

{Sentiment Analysis:}
The Natural Language Toolkit's (NLTK) VADER sentiment analysis library was used to calculate the tokenized sentiment for each tweet. Additionally, Google's BERT language model \cite{devlin2018bert} was used to calculate a semantic-based sentiment score for the returned tweets. Due to BERT's long computation time, BERT sentiment was calculated for the 100 most recent tweets. 

{Tweet Distribution:}
The tweet distribution is captured by post-processing the tweets and determining: (a) the number of tweets posted by the company / total tweets, (b) the number of tweets containing the company's username / total tweets, (c) the number of tweets containing the company's website / total tweets, (d) the number of tweets replying to a company tweet / total tweets, (e) the number of tweets replying to a company tweet / total tweets, and finally (f) the total number of tweets in the given nine-month period, capped at 5,000.  

{Tweet Contents:}
Regex and string matching is used to determine the number of hashtags, mentions, links, and emojis in each tweet. The number of distinct languages found across collected tweets, along with the detected topic as described in Section \ref{topic_modeling_setup} are used to capture the diversity of content and hype about major company events. 

{Linguistic Features:}
Natural language processing is performed to calculate metrics on the structure of the tweets, such as the tweet length, and character, token, and punctuation counts. Deep linguistic features were calculated to uncover semantic and syntactic patterns in the text. The Spacy library \cite{spacy2} was used for calculating grammatical features such as part-of-speech (POS) distribution, and PassivePy was used to calculate the number of times a passive voice was used in collected tweets. 

{Language Complexity:}
A high language complexity was hypothesized to  correlate to lower success as social media users typically communicate using colloquial language and could be dissuaded by complicated messages. The Flesch Reading-Ease score \cite{farr1951simplification} is used alongside the readability results capturing (a) long words (>7 letters), and (b) complex words, (>3 syllables) to capture tweet complexity. 

\subsection{Topic Modeling} \label{topic_modeling_setup}
A topic classification model was built to determine whether or not a news article, google search result, or tweet is about (a) funding events, (b) merger and acquisitions (c) geographical expansions, (d) new product launches, (e) awards received, and (f) management changes. These six topics are chosen as key milestones typical for companies experiencing growth and expansion, typical of successful companies. 

To train the classification model, a dataset of 3196 news article headlines from Crunchbase was collected and manually labeled as one of the above six topics, or other. Five different models are trained and evaluated: (a) Logistic Regression, (b) Random Forest, (c) Naive Bayes, (d) Support Vector Machine, and (e) XGBoost, with XGBoost selected as the dominant model. 

\subsection{Prediction Models} 
A total of ten ML classification models are implemented in this paper: K-Nearest Neighbors (KNN), Support Vector Machine (SVM), Logistic Regression (LR), Naive Bayes (NB), Decision Tree (DT), Random Forest (RF), AdaBoost (SB), Gradient Boost (AB), XGBoost (XGB), and CatBoost (CB).

The training and test datasets were loaded in from csv files, containing \traindatapts datapoints and \testdatapts datapoints respectively. There was a higher portion of negative datapoints so the dataset was balanced by upsampling the positive class to reduce bias in the prediction model. 

Categorical features such as industry categories (Ex. Software, Health Care, Financial Services) are converted to their respective indices since all of the models except CatBoost require numerical features for prediction. For the CatBoost model, the categorical feature list is passed in along with the training data as input when fitting the model.  

Next, the numerical features are normalized using the MinMaxScaler to ensure the model isn't favoring specific features over others.
\section{Experimental Setup} \label{experimental_setup}
The primary goal of this thesis is to predict startup success, measured by the raising of a funding round from investors. The process of evaluating the best funding prediction model was fourfold; First, ten different machine learning models are trained and evaluated to determine the highest performer. Next, the feature set is analyzed to determine the best collection of features that yields the highest results. Next, the best model along with the optimal feature set is used to compare how well the model performs on predicting funding events at different ranges into the future. Finally, the dataset is limited to only successful companies and evaluated on how well we could predict a second round of funding for companies that had already raised an initial round.  

A total of \anycomp companies are used in this section to generate \anydatapts data points, approximately 2 per company. These each represent a specific point in time during the company's history; more specifically, January 1st of a given year. Each data point contains all the features extracted from Crunchbase, Google, and Twitter and these are used to predict whether or not that company would raise a round of funding within a specified time horizon. 

\subsection{Model Performance}
A total of ten different prediction models are evaluated to determine the best predictor for company success, measured by the highest F1 score for the positive (raised funding) class. A total of 171 features are generated for each data point, representing general company data, funding events, news articles, Google search results, and Twitter social media activity for the target company. 

\subsection{Year Range Comparison}
Prior research in startup success prediction focused on predicting funding within a predetermined time horizon. This paper advances this research by exploring how well a model performs on predicting funding variable years into the future.

The same data points, each representing a company at a fixed point in time, are used in all forecasting instances. However, the y\_value, whether or not the company raised a funding round, is changed to reflect whether or not the company raised money in the upcoming 1-5 years. 

\subsection{Probability Cutoff}
By default, the Catboost boosting algorithm classifies data into the positive class if it's assigned a probability score of 50\% or greater for being in that class. In practice, an investor will only be able to fund a very small number of startups, so they're primarily interested in viewing only the top companies with a very high likelihood of success. Consequently, high precision is typically preferred at the expense of recall, and this ratio is adjusted by setting the probability cutoff to optimize the \(F_{0.1}\) score.

\subsection{Additional Funding Rounds}
The state-of-the-art prediction model found in our literature search \cite{sharchilev_web-based_2018}, predicts whether or not a company will raise a Series A or higher round of funding within the next 12 months. The authors chose angel and seed rounds as triggers to be included in the dataset, and consequently, all included companies have already raised a funding round. To measure our model's performance against this existing work, the data is filtered to only include companies that had raised an angel or seed round of investment. The CatBoost boosting method is trained on the finalized feature set using 0.5 and 0.75 cutoffs, and the precision, recall, and \(F_{0.1}\) scores are compared against the WBSSP\cite{sharchilev_web-based_2018} and FPAWI\cite{garkavenko_2022}. 
\section{Results}
This section reports on the performance of the machine learning models used in this paper, concluding with a comparison to prior work in this field.

\subsection{Model Performance}
Scores for each one of the ten prediction methods used can be viewed in Table \ref{tab:prediction-models}. These are trained on the entire feature set of 161 unique numerical fields, and the CatBoost model was trained on an additional 10 categorical fields. The goal is to predict whether or not the company will raise funding 3 years into the future given a starting date. CatBoost ensemble algorithm achieved the highest precision and F1 scores of 0.663 and 0.736, confirming previous findings \cite{garkavenko_2022} and is used in the rest of the analysis on this paper.

\begin{table*}
\caption{
    \label{tab:prediction-models} Prediction Method Performance
}

\begin{tabular}{ | l | c | c | c | c | c | c |} 
    \hline
    \centering Model & Overall F1 &
    Funding Precision & Funding Recall & Funding F1 \\
    \hline
    Naive Bayes & 0.6197 & 0.3780 & 0.5212 & 0.4382 \\
    K Nearest Neighbor & 0.6461 & 0.4009 & 0.6500 & 0.4960 \\
    Decision Tree Classifier & 0.6815 & 0.4737 & 0.5695 & 0.5172 \\
    Random Forest & 0.7692 & 0.6392 & 0.6406 & 0.6399 \\
    AdaBoost & 0.7624 & 0.5452 & 0.8227 & 0.6558 \\
    Support Vector Classifier & 0.7654 & 0.5579 & 0.7955 & 0.6558 \\
    XGBoost & 0.7755 & 0.5935 & 0.7500 & 0.6627 \\
    Gradient Boost & 0.7673 & 0.5513 & \textbf{0.8312} & 0.6628 \\
    Logistic Regression & 0.7718 & 0.5694 & 0.7955 & 0.6637 \\
    CatBoost & \textbf{0.8250} & \textbf{0.6634} & 0.8273 & \textbf{0.7363} \\		
    \hline
\end{tabular}
\end{table*}

\subsection{Feature Evaluation}
\begin{figure}
  \includegraphics[width=.45\textwidth]{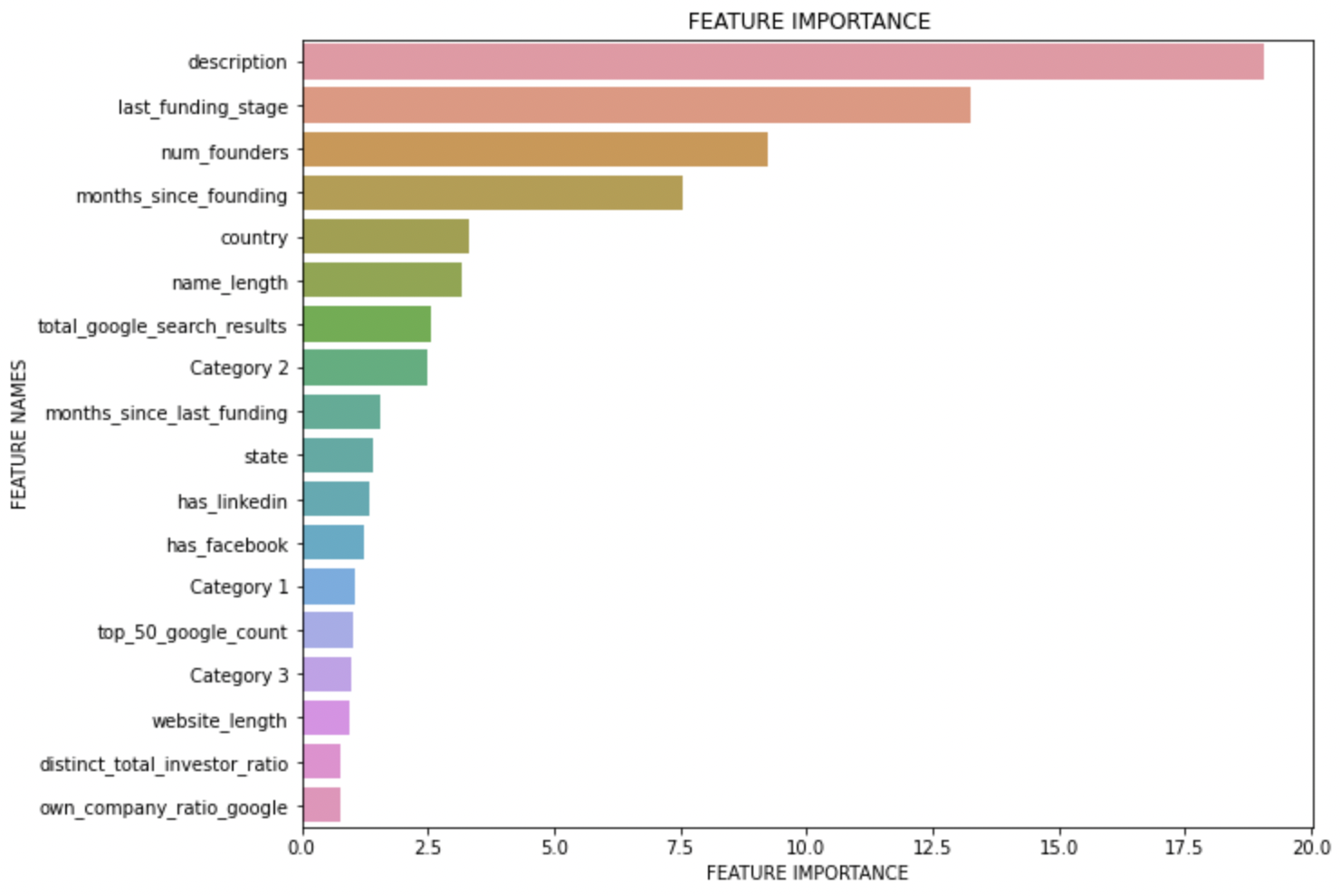}
  \caption{Top 18 Catboost Features}
  \label{fig:top-18-catboost}
\end{figure}
A list of the top 18 features by Catboost training weight can be viewed in Figure \ref{fig:top-18-catboost}. A model trained using only the aforementioned 18 features achieves an F1 score of 0.718, scoring very close to the 0.736 using the entire feature set. The company description, last funding stage, number of founders, and months since founding make the biggest impact on model performance. 

\subsection{Year Range Comparison}
The distribution of datapoints in the positive class varies greatly between the different date ranges (Table \ref{tab:year-range}). Only 10.9\% of startups in the evaluation dataset raised funding within a year, compared to 22.6\% that raised funding looking 5 years into the future, a 107\% increase. This means there are less positive examples in the training set for the model to train on, and the CatBoost method achieved an F1 score of 0.5282 with the limited training data. 

\begin{table}

\caption{
    \label{tab:year-range} Predicting Funding Variably into the Future 
}

\begin{tabular}{ | l | c | c |  c | c | c |} 
    \hline
    Range & Precision & Recall & F1 & Received Funding \\
    \hline
    1 year & 0.6224 & 0.4587 & 0.5282 & 10.9\% \\
    2 years & 0.6379 & 0.7349 & 0.6830 & 17.1\% \\
    3 years & 0.6634 & 0.8273 & 0.7363 & 22.0\% \\
    4 years & 0.6847 & 0.8622 & 0.7633 & 22.5\% \\
    5 years & \textbf{0.7042} & \textbf{0.8660} & \textbf{0.7768} & 22.6\% \\
    \hline
\end{tabular}
\end{table}

The model performance improves as the year range increases, scoring 0.6830, 0.7363, 0.7633, and 0.7768 looking 2, 3, 4, and 5 years into the future. The model experiences the biggest jumps in improvement between year 1 and 2 (29.3\%) and year 2 and year 3 (7.8\%). After year 3, the rate of improvement drops to 3.7\%. Additionally, while the number of companies raising funding more than doubles between year 1 and year 3, there's only a 2.7\% increase between year 3 and 5. Since investors want a return on investment as soon as possible, the paper primarily uses the 3 year range to balance real-world usability with model performance. 

\subsection{Probability Cutoff}
Classification scores for cutoff thresholds ranging from 0.5 to 0.95 are calculated in steps of 0.05 (see Table \ref{tab:probability-performance}). Additionally, performance is measured with the cutoff set at 0.97 and 0.99 to capture the highest bias towards precision over recall. Increasing the cutoff threshold for being classified in the positive class steadily improved precision, starting at 0.6634 and reaching the maximum precision of 1.0 at the 0.97 cutoff threshold. In contrast, recall steadily dropped from 0.8273 to 0.0061, meaning fewer companies fit the steep cutoff threshold to be classified as successful. The best $F_1$ score is achieved at the 0.55 cutoff, where precision and recall are almost equally favored. Setting beta at 0.1 heavily favors precision over recall, and the best $F_{0.1}$ score is achieved at the 0.9 cutoff with a score of 0.9265. At the 0.9 cutoff threshold, a precision of 0.9505 and a recall of 0.2621 are achieved. However, the dataset isn't large enough to fully generalize a 95\% precision score across all startups, industries, and countries.
\begin{table}
\caption{
    \label{tab:probability-performance} Performance by Cutoff Threshold
}

\begin{tabular}{ | c | c | c | c | c |} 
    \hline
    Cutoff & Precision & Recall & $F_{1}$ & $F_{0.1}$ \\
    \hline
    0.99 & \textbf{1.0000} & 0.0061 & 0.0120 & 0.3811 \\
    0.97 & \textbf{1.0000} & 0.0424 & 0.0814 & 0.8173 \\
    0.95 & 0.9512 & 0.1182 & 0.2102 & 0.8892 \\
    0.90 & 0.9505 & 0.2621 & 0.4109 & \textbf{0.9265} \\
    0.85 & 0.9039 & 0.3848 & 0.5399 & 0.8920 \\
    0.80 & 0.8676 & 0.4864 & 0.6233 & 0.8609 \\
    0.75 & 0.8283 & 0.5773 & 0.6804 & 0.8247 \\
    0.70 & 0.7792 & 0.6364 & 0.7006 & 0.7775 \\
    0.65 & 0.7562 & 0.6955 & 0.7245 & 0.7555 \\
    0.60 & 0.7269 & 0.7500 & 0.7383 & 0.7271 \\
    0.55 & 0.7009 & 0.7955 & \textbf{0.7452} & 0.7018 \\
    0.50 & 0.6634 & \textbf{0.8273} & 0.7363 & 0.6647 \\
    \hline
\end{tabular}
\end{table}

\subsection{Additional Funding Rounds}
To complete the evaluation of our results, our best CatBoost boosting model using the entire feature set is compared to prior work in the startup prediction space.

Sharchilev et al. achieved state-of-the-art performance in 2018 with their Gradient Boosting Decision Tree trained on Crunchbase, LinkedIn, and internet data collected from the Yandex Search Engine \cite{sharchilev_web-based_2018}. Garkavenko et al. further advanced the field of startup success prediction with their 2022 paper using online, publicly available data from the startup's own website, the Google Search API, and Twitter to collect 17 distinct features for each startup \cite{garkavenko_2022}. Although the research papers didn't share the datasets used, care was taken to replicate the experimental setup as best as possible. In particular, an angel or seed round of investment is a required trigger to be considered, and the goal is to predict a Series A or later round of funding within one year of a provided date.

The results are displayed in Table \ref{tab:related-work-comparison}. Our default CatBoost model trained on the entire feature set achieves state-of-the-art performance with precision and $F_{0.1}$ scores of 0.656 and 0.655 respectively. The WBSSP and FPAWI models achieved $F_{0.1}$ scores of 0.383 and 0.531 at the same task of identifying Series A or later funding rounds. Furthermore, our best Catboost model adjusted with a cutoff threshold of 0.75 scores 0.744 and 0.730 on precision and $F_{0.1}$, surpassing all other models. While increasing the cutoff threshold to 0.90 further increases the precision and $F_{0.1}$ scores to 1.0 and 0.869, this isn't used as the final model as the dataset isn't large enough to dismiss overfitting in the evaluation set.  
\begin{table}
\setlength{\tabcolsep}{8pt} 

\caption{
    \label{tab:related-work-comparison} Comparison to results reported in \cite{garkavenko_2022} and \cite{sharchilev_web-based_2018}
}

\begin{tabular}{ | l | c | c | c | c | c | } 
    \hline
    Model & Datapoints & Features & Precision & $F_{0.1}$ \\
    \hline
    $CB_{0.75}$* & 1,888 & 18 & \textbf{0.790} & \textbf{0.774} \\
    $CB_{0.75}$ & 1,888 & 171 & 0.744 & 0.730 \\
    $CB_{0.50}$ & 1,888 & 171 & 0.656 & 0.655\\
    FPAWI & 33,165  & 17   & 0.640 & 0.531 \\
    WBSSP & 15,128 & 49   & 0.626 & 0.383 \\
    \hline
\end{tabular}
\end{table}
\section{Conclusion}
This paper explores if startup success prediction with additional features and textual analysis can advance to state of the art. More specifically, can we predict whether or not a company can raise a round of funding within an allotted time frame?

Information on general company data, previous funding rounds, published news articles, Google Search results, and Twitter social media activity is collected for \totalcomp companies on Crunchbase. These companies were founded between 2007 and 2021 and 4,086 (41.5\%) had raised at least one funding round. A total of 171 features are collected for every datapoint, comprised of 161 numerical and 10 categorical features. The 18 best performing features are listed in Figure \ref{fig:top-18-catboost}, on their own achieving an F1 score of 0.718, while all 171 features together perform at an F1 score of 0.736.

The CatBoost ensemble method achieves the best performance with precision, recall, and F1 scores of 0.663, 0.827, and 0.736 respectively at the 3-year prediction task. The same ensemble method achieves F1 scores of 0.528, 0.683, 0.736, 0.763, and 0.777 when tasked with predicting a funding round one to five years into the future. By adjusting the cutoff threshold to 0.9 to favor precision, the CatBoost model reaches a maximum $F_{0.1}$ score of 0.927. The final objective is predicting whether or not a startup that had already raised an angel or seed round of funding would raise another round within one year of the provided date. Our best CatBoost model trained on all 171 features and a 0.75 cutoff threshold achieves precision and \(F_{0.1}\) scores of 0.744 and 0.730 significantly improving on results from Table \ref{tab:related-work-comparison}.

\subsection{Reflection}
It's important to note that while several features have been found to highly correlate to startup success, they don't capture the internal workings of a company, only outward signals. The number of founders, the company description, and the complexity of tweets are indicators of the underlying workings of a company, but can be artificially mimicked to deceive the prediction model. Simply finding a random person to join your co-founder team and then posting much content on social media, while following what successful companies are doing, fails to address the key components that make a startup succeed. We advise using prediction models only as the first step to screening startups, and then doing due diligence to confirm that a business is worth investing in.  

\subsection{NLP Performance}
Many of the public sources used in this study were textual and thus NLP was used to derive features which themselves could be textual or categorical. While the company description achieved the best individual feature score, not many of the other NLP-heavy features, notably Twitter analysis, were part of the top 18. However, these features clearly still contributed to the overall prediction strength since the F1 performance on all 171 features is higher than the top 18. 

\subsection{Future Work}
Our work helps improve startup success prediction and increases understanding of what characteristics indicate a successful startup. However, there is more work that can be done in this area to enhance the dataset and provide individuals with actionable insights using the prediction model. 

\subsubsection{Data Collection}
We are limited in the number of companies contained in our dataset due to the limited duration of this research and the API limits set by the Twitter and Google Search APIs. Future work utilizing a higher volume of companies would be better equipped to avoid over-fitting to the training set. The 300 million startups created each year are made up of a diverse range of industries, geographic regions, and growth profiles, and a larger dataset would better capture the various trends across the startup space.

\subsubsection{Founder Profiles}
Previous work into startup prediction has found correlation between the founders' background and the success of their company. Information contained on Crunchbase and LinkedIn can be used to collect data on a founder's educational history, previous ventures, and social networks. These additional fields may provide clues into the individuals' professional network, as people often say one's net worth is their network. Additionally, a founder with a history of successful ventures is more likely to start another successful company than a first-time founder. 

\subsubsection{Company Size}
Small, early-stage companies with less than a dozen employees need a lot less funding than huge businesses overseeing hundreds to thousands of people. Consequently, the amount of funding, and the extent to which funding is needed, varies significantly across the stage the company is at and how many people need to be paid. While a partial view of the stage of a business is contained in the months since founding and the last investment stage features, taking into account the company size will help paint a more wholistic view of the business. Additionally, normalizing metrics by company size can help alleviate the huge disparity between early to later-stage startups, as the number of likes received on social media will vary significantly between new ventures and businesses with thousands of employees and millions of dollars in revenue. 

\subsubsection{Performance Across Sectors}
This paper focused on predicting whether or not a company would raise a funding round, but didn't dive into what companies it was particularly good or bad at predicting. In particular, it would be interesting to see how well the prediction model performs across different industries: does it favor one sector such as Software or Biotech over others. Additionally, further analysis is needed to determine how well the model fares across different states and countries.

\subsubsection{Large Language Models}
LLMs have recently revolutionized many aspects of scientific and linguistic analysis. They can be used for more feature analysis and derivation of insight. Of particular interest would be a new LLM approach to the ``wisdom of crowd'' concept whereby the LLM training corpus can be mined for features. 



\bibliographystyle{ACM-Reference-Format}
\bibliography{main}


\begin{thebibliography}{21}


\ifx \showCODEN    \undefined \def \showCODEN     #1{\unskip}     \fi
\ifx \showDOI      \undefined \def \showDOI       #1{#1}\fi
\ifx \showISBNx    \undefined \def \showISBNx     #1{\unskip}     \fi
\ifx \showISBNxiii \undefined \def \showISBNxiii  #1{\unskip}     \fi
\ifx \showISSN     \undefined \def \showISSN      #1{\unskip}     \fi
\ifx \showLCCN     \undefined \def \showLCCN      #1{\unskip}     \fi
\ifx \shownote     \undefined \def \shownote      #1{#1}          \fi
\ifx \showarticletitle \undefined \def \showarticletitle #1{#1}   \fi
\ifx \showURL      \undefined \def \showURL       {\relax}        \fi
\providecommand\bibfield[2]{#2}
\providecommand\bibinfo[2]{#2}
\providecommand\natexlab[1]{#1}
\providecommand\showeprint[2][]{arXiv:#2}

\bibitem[cru(2022)]%
        {crunchbase_csv_export}
 \bibinfo{year}{2022}\natexlab{}.
\newblock \bibinfo{title}{{Crunchbase Daily CSV Export}}.
\newblock
\newblock
\newblock
\shownote{\url{https://data.crunchbase.com/docs/daily-csv-export}}.


\bibitem[Antosiuk(2021)]%
        {antosiuk_predicting_2021}
\bibfield{author}{\bibinfo{person}{Piotr Antosiuk}.}
  \bibinfo{year}{2021}\natexlab{}.
\newblock \bibinfo{title}{Predicting startup success with machine learning
  methods}.
\newblock
\newblock
\urldef\tempurl%
\url{https://repo.pw.edu.pl/info/master/WUTadfafd4ff4284265b3820d0743f24cba/}
\showURL{%
\tempurl}


\bibitem[Antretter et~al\mbox{.}(2019)]%
        {antretter_predicting_2019}
\bibfield{author}{\bibinfo{person}{Torben Antretter}, \bibinfo{person}{Ivo
  Blohm}, \bibinfo{person}{Dietmar Grichnik}, {and} \bibinfo{person}{Joakim
  Wincent}.} \bibinfo{year}{2019}\natexlab{}.
\newblock \showarticletitle{Predicting new venture survival: A Twitter-based
  machine learning approach to measuring online legitimacy}.
\newblock  (\bibinfo{year}{2019}).
\newblock
\showISSN{23526734}
\urldef\tempurl%
\url{https://linkinghub.elsevier.com/retrieve/pii/S2352673418301197}
\showURL{%
\tempurl}


\bibitem[Dellermann et~al\mbox{.}(2017)]%
        {dellermann_2017}
\bibfield{author}{\bibinfo{person}{Dominik Dellermann},
  \bibinfo{person}{Nikolaus Lipusch}, \bibinfo{person}{Philipp Ebel},
  \bibinfo{person}{Karl~Michael Popp}, {and} \bibinfo{person}{Jan~Marco
  Leimeister}.} \bibinfo{year}{2017}\natexlab{}.
\newblock \showarticletitle{Finding the Unicorn: Predicting Early Stage Startup
  Success Through a Hybrid Intelligence Method}.
\newblock  (\bibinfo{year}{2017}).
\newblock
\showISSN{1556-5068}
\urldef\tempurl%
\url{https://doi.org/10.2139/ssrn.3159123}
\showDOI{\tempurl}


\bibitem[Devlin et~al\mbox{.}(2018)]%
        {devlin2018bert}
\bibfield{author}{\bibinfo{person}{Jacob Devlin}, \bibinfo{person}{Ming-Wei
  Chang}, \bibinfo{person}{Kenton Lee}, {and} \bibinfo{person}{Kristina
  Toutanova}.} \bibinfo{year}{2018}\natexlab{}.
\newblock \showarticletitle{Bert: Pre-training of deep bidirectional
  transformers for language understanding}.
\newblock \bibinfo{journal}{\emph{arXiv preprint arXiv:1810.04805}}
  (\bibinfo{year}{2018}).
\newblock


\bibitem[Farr et~al\mbox{.}(1951)]%
        {farr1951simplification}
\bibfield{author}{\bibinfo{person}{James~N Farr}, \bibinfo{person}{James~J
  Jenkins}, {and} \bibinfo{person}{Donald~G Paterson}.}
  \bibinfo{year}{1951}\natexlab{}.
\newblock \showarticletitle{Simplification of Flesch reading ease formula.}
\newblock \bibinfo{journal}{\emph{Journal of applied psychology}}
  \bibinfo{volume}{35}, \bibinfo{number}{5} (\bibinfo{year}{1951}),
  \bibinfo{pages}{333}.
\newblock


\bibitem[Garkavenko et~al\mbox{.}(2022)]%
        {garkavenko_2022}
\bibfield{author}{\bibinfo{person}{Mariia Garkavenko}, \bibinfo{person}{Eric
  Gaussier}, \bibinfo{person}{Hamid Mirisaee}, \bibinfo{person}{Cédric
  Lagnier}, {and} \bibinfo{person}{Agnès Guerraz}.}
  \bibinfo{year}{2022}\natexlab{}.
\newblock \bibinfo{title}{Where Do You Want To Invest? Predicting Startup
  Funding From Freely, Publicly Available Web Information}.
\newblock
\newblock
\showeprint[arxiv]{2204.06479}~[cs.CE]


\bibitem[Gavrilenko(2022)]%
        {gavrilenko_2022}
\bibfield{author}{\bibinfo{person}{Emily Gavrilenko}.}
  \bibinfo{year}{2022}\natexlab{}.
\newblock \bibinfo{title}{Predicting Startup Success using Publicly Available
  Data}.
\newblock
\newblock
\urldef\tempurl%
\url{https://digitalcommons.calpoly.edu/theses/2652/}
\showURL{%
\tempurl}


\bibitem[Honnibal and Montani(2017)]%
        {spacy2}
\bibfield{author}{\bibinfo{person}{Matthew Honnibal} {and}
  \bibinfo{person}{Ines Montani}.} \bibinfo{year}{2017}\natexlab{}.
\newblock \bibinfo{title}{{spaCy 2}: Natural language understanding with
  {B}loom embeddings, convolutional neural networks and incremental parsing}.
  (\bibinfo{year}{2017}).
\newblock


\bibitem[{ISP}(2020)]%
        {isp_startups_created}
\bibfield{author}{\bibinfo{person}{{NetShop} {ISP}}.}
  \bibinfo{year}{2020}\natexlab{}.
\newblock \bibinfo{booktitle}{\emph{How Many Tech Startups Are Created Each
  Year?}}
\newblock
\urldef\tempurl%
\url{https://netshopisp.medium.com/how-many-tech-startups-are-created-each-year-27539d0a4c48}
\showURL{%
\tempurl}


\bibitem[Kampakis and Adamides({[n.\,d.]})]%
        {twitter_football}
\bibfield{author}{\bibinfo{person}{Stylianos Kampakis} {and}
  \bibinfo{person}{Andreas Adamides}.} \bibinfo{year}{[n.\,d.]}\natexlab{}.
\newblock \bibinfo{title}{Using Twitter to predict football outcomes}.
\newblock
\newblock
\urldef\tempurl%
\url{https://arxiv.org/pdf/1411.1243.pdf}
\showURL{%
\tempurl}


\bibitem[Khan et~al\mbox{.}(2021)]%
        {twitter_elections_2021}
\bibfield{author}{\bibinfo{person}{Asif Khan}, \bibinfo{person}{Huaping Zhang},
  \bibinfo{person}{Nada Boudjellal}, \bibinfo{person}{Arshad Ahmad},
  \bibinfo{person}{Jianyun Shang}, \bibinfo{person}{Lin Dai}, {and}
  \bibinfo{person}{Bashir Hayat}.} \bibinfo{year}{2021}\natexlab{}.
\newblock \bibinfo{title}{Election prediction on twitter: A systematic mapping
  study}.
\newblock
\newblock
\urldef\tempurl%
\url{https://www.hindawi.com/journals/complexity/2021/5565434/}
\showURL{%
\tempurl}


\bibitem[Murray(2021)]%
        {murray_overcoming_2021}
\bibfield{author}{\bibinfo{person}{Bryce Murray}.}
  \bibinfo{year}{2021}\natexlab{}.
\newblock \showarticletitle{Overcoming {AI} bias in predicting startup
  success}.
\newblock  (\bibinfo{year}{2021}).
\newblock
\urldef\tempurl%
\url{https://towardsdatascience.com/overcoming-ai-faults-in-predicting-startup-success-768985e6e289}
\showURL{%
\tempurl}


\bibitem[Sharchilev et~al\mbox{.}(2018)]%
        {sharchilev_web-based_2018}
\bibfield{author}{\bibinfo{person}{Boris Sharchilev}, \bibinfo{person}{Michael
  Roizner}, \bibinfo{person}{Andrey Rumyantsev}, \bibinfo{person}{Denis
  Ozornin}, \bibinfo{person}{Pavel Serdyukov}, {and} \bibinfo{person}{Maarten
  de Rijke}.} \bibinfo{year}{2018}\natexlab{}.
\newblock \showarticletitle{Web-based Startup Success Prediction}. In
  \bibinfo{booktitle}{\emph{Proceedings of the 27th {ACM} International
  Conference on Information and Knowledge Management}} (New York, {NY}, {USA},
  2018-10-17). \bibinfo{publisher}{Association for Computing Machinery}.
\newblock
\showISBNx{978-1-4503-6014-2}
\urldef\tempurl%
\url{https://doi.org/10.1145/3269206.3272011}
\showDOI{\tempurl}


\bibitem[Singh et~al\mbox{.}(2017)]%
        {twitter_disaster_classification}
\bibfield{author}{\bibinfo{person}{Jyoti Singh}, \bibinfo{person}{Yogesh
  Dwivedi}, \bibinfo{person}{Nripendra Rana}, \bibinfo{person}{Abhinav Kumar},
  {and} \bibinfo{person}{Kawaljeet Kapoor}.} \bibinfo{year}{2017}\natexlab{}.
\newblock \bibinfo{title}{Event classification and location prediction from
  tweets during disasters}.
\newblock
\newblock
\urldef\tempurl%
\url{https://d-nb.info/1132940923/34}
\showURL{%
\tempurl}


\bibitem[Soler et~al\mbox{.}(2012)]%
        {twitter_spanish_elections_2012}
\bibfield{author}{\bibinfo{person}{Juan Soler}, \bibinfo{person}{Fernando
  Cuartero}, {and} \bibinfo{person}{Manuel Roblizo}.}
  \bibinfo{year}{2012}\natexlab{}.
\newblock \bibinfo{title}{Twitter as a Tool for Predicting Elections Results}.
\newblock
\newblock
\urldef\tempurl%
\url{https://ieeexplore.ieee.org/abstract/document/6425594}
\showURL{%
\tempurl}


\bibitem[Stuart and Abetti(1987)]%
        {stuart_abetti_1987}
\bibfield{author}{\bibinfo{person}{Robert Stuart} {and} \bibinfo{person}{Pier
  Abetti}.} \bibinfo{year}{1987}\natexlab{}.
\newblock \bibinfo{title}{Start-up ventures: Towards the prediction of initial
  success}.
\newblock
\newblock
\urldef\tempurl%
\url{https://papers.ssrn.com/sol3/papers.cfm?abstract_id=1504468}
\showURL{%
\tempurl}


\bibitem[Tumasjan et~al\mbox{.}(2021)]%
        {tumasjan_twitter_2021}
\bibfield{author}{\bibinfo{person}{Andranik Tumasjan}, \bibinfo{person}{Reiner
  Braun}, {and} \bibinfo{person}{Barbara Stolz}.}
  \bibinfo{year}{2021}\natexlab{}.
\newblock \showarticletitle{Twitter sentiment as a weak signal in venture
  capital financing}.
\newblock  (\bibinfo{year}{2021}).
\newblock
\urldef\tempurl%
\url{https://linkinghub.elsevier.com/retrieve/pii/S0883902620306704}
\showURL{%
\tempurl}


\bibitem[Xiang et~al\mbox{.}(2012)]%
        {xiang_2012}
\bibfield{author}{\bibinfo{person}{Guang Xiang}, \bibinfo{person}{Zeyu Zheng},
  \bibinfo{person}{Miaomiao Wen}, \bibinfo{person}{Jason Hong},
  \bibinfo{person}{Carolyn Rose}, {and} \bibinfo{person}{Chao Liu}.}
  \bibinfo{year}{2012}\natexlab{}.
\newblock \showarticletitle{A Supervised Approach to Predict Company
  Acquisition With Factual and Topic Features Using Proﬁles and News Articles
  on {TechCrunch}}.
\newblock  (\bibinfo{year}{2012}).
\newblock


\bibitem[Ünal and Ceasu(2019)]%
        {unal_machine_2019}
\bibfield{author}{\bibinfo{person}{Cemre Ünal} {and} \bibinfo{person}{Ioana
  Ceasu}.} \bibinfo{year}{2019}\natexlab{}.
\newblock \bibinfo{booktitle}{\emph{A Machine Learning Approach Towards Startup
  Success Prediction}}.
\newblock
\urldef\tempurl%
\url{https://www.econstor.eu/handle/10419/230798}
\showURL{%
\tempurl}


\bibitem[Żbikowski and Antosiuk(2021)]%
        {zbikowski_machine_2021}
\bibfield{author}{\bibinfo{person}{Kamil Żbikowski} {and}
  \bibinfo{person}{Piotr Antosiuk}.} \bibinfo{year}{2021}\natexlab{}.
\newblock \showarticletitle{A machine learning, bias-free approach for
  predicting business success using Crunchbase data}.
\newblock \bibinfo{journal}{\emph{Information Processing \& Management}}
  (\bibinfo{year}{2021}).
\newblock
\urldef\tempurl%
\url{https://www.sciencedirect.com/science/article/pii/S0306457321000595}
\showURL{%
\tempurl}


\end{thebibliography}


\end{document}